\pdfoutput=1
\documentclass[11pt]{article}

\usepackage{ACL2023}

\usepackage{times}
\usepackage{latexsym}

\usepackage[T1]{fontenc}
\usepackage[utf8]{inputenc}

\usepackage{microtype}

\usepackage{inconsolata}
\usepackage{latexsym}
\usepackage{amsmath}
\usepackage{multicol}
\usepackage{multirow}
\usepackage{algorithm}
\usepackage{algpseudocode}
\usepackage{latexsym}
\usepackage{graphicx}
\usepackage{makecell}
\usepackage{booktabs}
\usepackage{pifont}
\usepackage{colortbl}
\usepackage{longtable}
\renewcommand{\Return}[1]{\State{\textbf{return} #1}}
\newenvironment{breakablealgorithm}
{% \begin{breakablealgorithm}
    \begin{center}
        \refstepcounter{algorithm}% New algorithm
        \hrule height.8pt depth0pt \kern2pt% \@fs@pre for \@fs@ruled
        \renewcommand{\caption}[2][\relax]{% Make a new \caption
            {\raggedright\textbf{\textbf{Algorithm}~\thealgorithm} ##2\par}%
            \ifx\relax##1\relax % #1 is \relax
            \addcontentsline{loa}{algorithm*}{\protect\numberline{\thealgorithm}##2}%
            \else % #1 is not \relax
            \addcontentsline{loa}{algorithm*}{\protect\numberline{\thealgorithm}##1}%
            \fi
            \kern2pt\hrule\kern2pt
        }
    }{% \end{breakablealgorithm}
        \kern2pt\hrule\relax% \@fs@post for \@fs@ruled
    \end{center}
}
\makeatother

\newcommand{\realred}[1]{\textcolor{red}{#1}}
\newcommand{\realblue}[1]{\textcolor{blue}{#1}}

\title{From the One, Judge of the Whole: Typed Entailment Graph Construction with Predicate Generation}

\author{Zhibin Chen$^{123}$\quad Yansong Feng$^{13}$\thanks{\ \ Corresponding author.}\quad Dongyan Zhao$^{123}$\\
$^1$Wangxuan Institute of Computer Technology, Peking University, China\\
$^2$Center for Data Science, Peking University, China\\
$^3$The MOE Key Laboratory of Computational Linguistics, Peking University, China\\
\texttt{\{czb-peking, fengyansong, zhaody\}@pku.edu.cn}}

\begin{document}
\maketitle
\begin{abstract}

Entailment Graphs (EGs) have been constructed based on extracted corpora as a strong and explainable form to indicate context-independent entailment relations in natural languages. However, EGs built by previous methods often suffer from the severe sparsity issues, due to limited corpora available and the long-tail phenomenon of predicate distributions. In this paper, we propose a multi-stage method, Typed Predicate-Entailment Graph Generator (TP-EGG), to tackle this problem. Given several seed predicates, TP-EGG builds the graphs by generating new predicates and detecting entailment relations among them. The generative nature of TP-EGG helps us leverage the recent advances from large pretrained language models (PLMs), while avoiding the reliance on carefully prepared corpora.
Experiments on benchmark datasets show that TP-EGG can generate high-quality and scale-controllable entailment graphs, achieving significant in-domain improvement over state-of-the-art EGs and boosting the performance of down-stream inference tasks\footnote{Our code is available at \url{https://github.com/ZacharyChenpk/TP-EGG}}.

\end{abstract}

\section{Introduction}

The entailment relation between textual predicates plays a critical role in natural language inference and natural language understanding tasks, including question answering~\citep{pathak2021scientific,mckenna-etal-2021-multivalent} and knowledge graph completion~\citep{yoshikawa2019combining,hosseini-etal-2019-duality,hosseini-etal-2021-open-domain}. To detect entailment relations, previous works pay attention to the Recognizing Textual Entailment (RTE) task, which takes a pair of sentences as input and predicts whether one sentence entails the other~\citep{bowman-etal-2015-large,he2021deberta,DBLP:journals/corr/abs-2009-09139}. Current RTE models perform well on RTE benchmarks, 
but most of them are lacking in explainability, as they make use of the black-box Language Models (LM) without providing any explainable clues.

\begin{figure}[t!]
    \centering
    \includegraphics[width=0.7\linewidth]{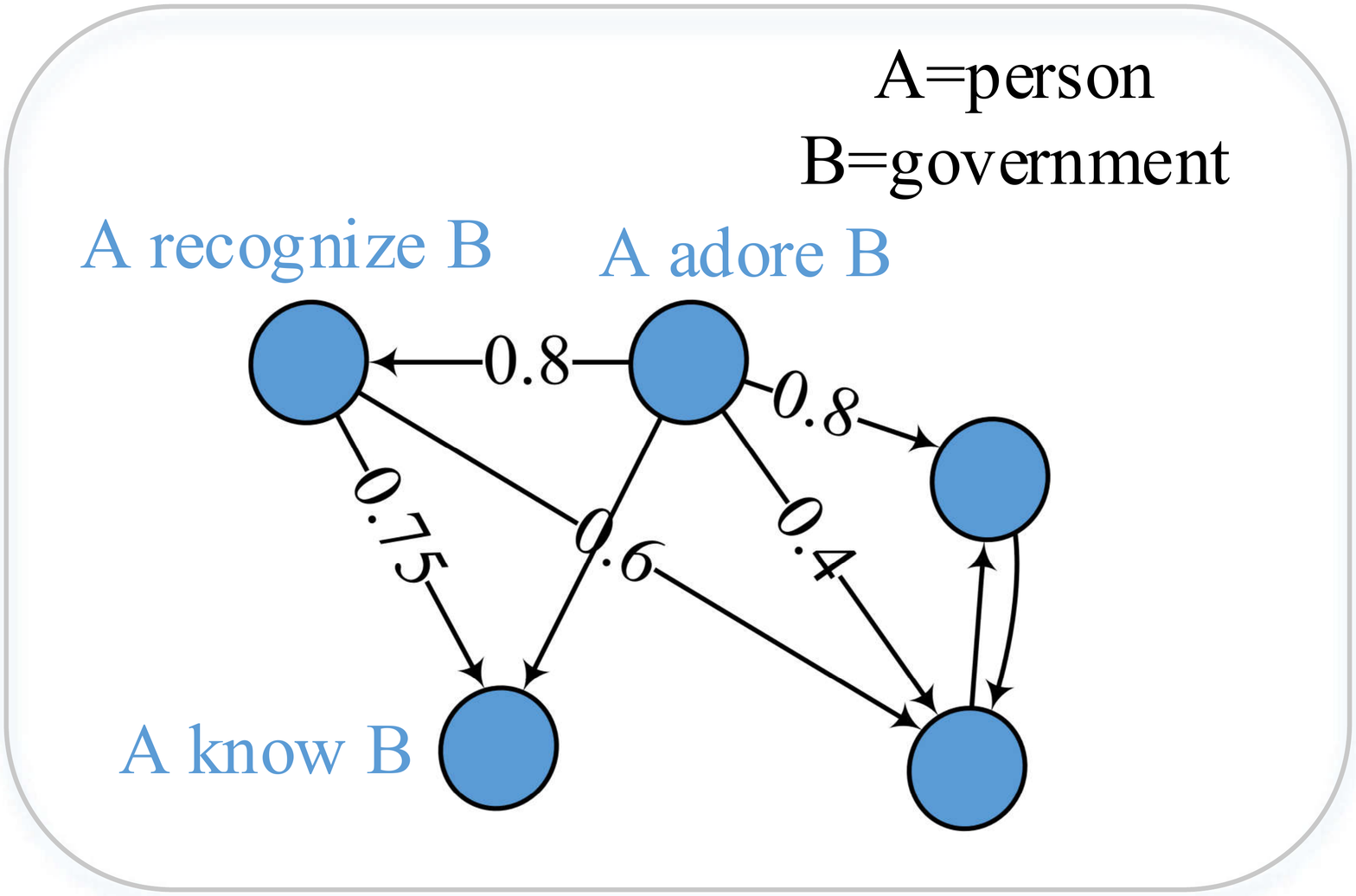}
    \caption{An example of typed entailment graph including several predicates, with argument types \emph{person} and \emph{government}.}
    \label{fig:eg}
\end{figure}

Recent works focus on learning the Entailment Graph (EG) structure, which organizes typed predicates in directional graphs with entailment relations as the edges~\citep{hosseini-etal-2018-learning,hosseini-etal-2019-duality,mckenna-etal-2021-multivalent}, as shown in Figure~\ref{fig:eg}. With the explicit graph structure containing predicates and their entailment relations, similar to Knowledge Graphs (KGs), using EGs becomes an explainable and context-independent way to represent the knowledge required in natural language inference and other NLP tasks.

\begin{figure*}[t!]
    \centering
    \includegraphics[width=0.9\linewidth]{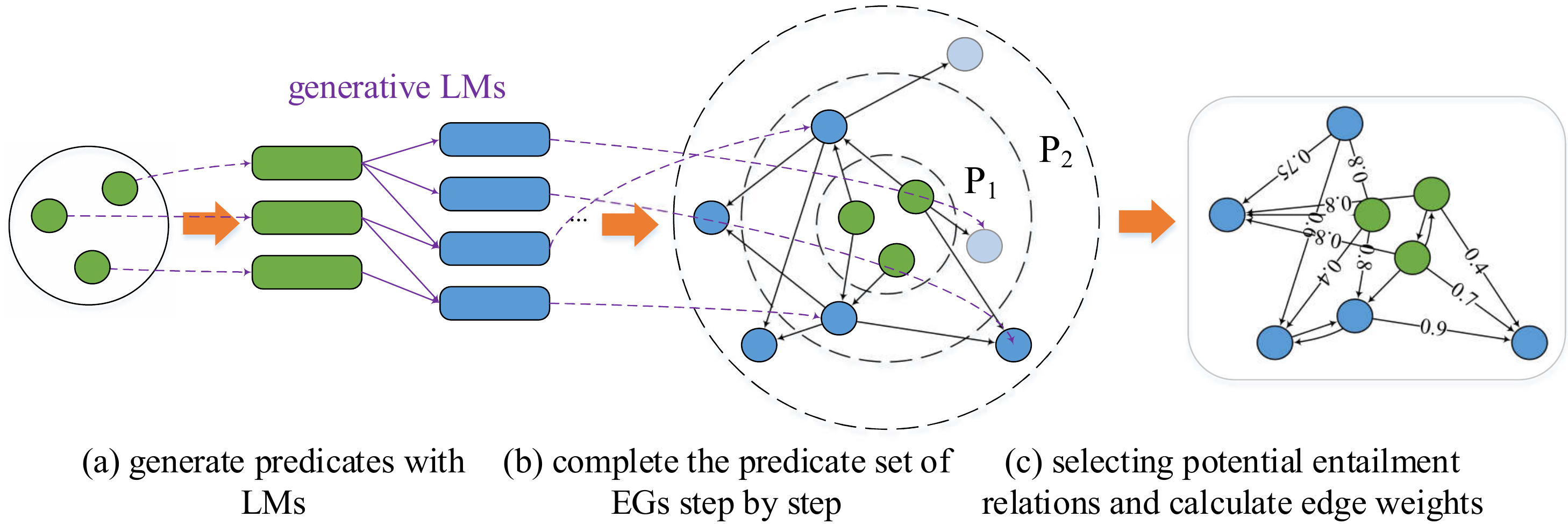}
    \caption{An illustration of our TP-EGG. Given three seed predicates, TP-EGG generates a graph with 8 predicates and 15 entailment relations. The circles represents different predicates, while the rounded rectangles is sentences in natural language. Seed predicates is in \textcolor[RGB]{112,173,71}{green}, and newly generated predicates is in \textcolor[RGB]{91,155,213}{blue}.}
    \label{fig:model}
\end{figure*}

Most existing EGs are constructed with the Distributional Inclusion Hypothesis (DIH), which suggests that all typical context features of a predicate $v$ can also occur with another predicate $w$ if $v$ entails $w$~\citep{geffet-dagan-2005-distributional}. Constructing EGs with DIH requires distributional co-occurrences of contextual features from large corpora to calculate the semantic similarity between predicates~\citep{szpektor-dagan-2008-learning,schoenmackers-etal-2010-learning}. However, the EGs constructed from large corpora often suffer from two different kinds of sparsity issues: \emph{the predicate sparsity} and \emph{the edge sparsity}. Existing corpora used for EG construction are mainly collected from specific resources~\citep{zhang-weld-2013-harvesting}, such as news articles. As a result, entailment relations could not be learned between those predicates that do not appear in the corpora, which leads to the \emph{predicate sparsity} issue. Meanwhile, if two predicates scarcely appear around similar contexts in the given corpora, the DIH could not indicate the potential entailment relationship between them. It leads to the \emph{edge sparsity} of EGs as the corresponding edges may be missing due the limited coverage of the corpora.

To tackle the sparsity issues, previous works pay attention to learning global graph structures to mine latent entailment relations and alleviate the edge sparsity~\citep{berant-etal-2011-global,berant-etal-2015-efficient,hosseini-etal-2018-learning,chen-etal-2022-entailment}, but predicate sparsity is still holding back the improvement of EGs. Solving predicate sparsity by simply scaling up the distributional feature extraction is impracticable, due to the long-tail phenomenon of predicate distribution~\citep{McKenna2022SmoothingEG}. 

The shortcomings of extractive methods come in quest for non-extraction way to overcome. Recent progress in deep generative LMs, including GPT-3~\citep{NEURIPS2020_1457c0d6} and T5~\cite{10.5555/3455716.3455856}, makes it possible to produce predicates and entailment relations by generative methods. Inspired by the Commonsense Transformer \citep{bosselut-etal-2019-comet}, we propose a novel generative multi-stage EG construction method, called Typed Predicate-Entailment Graph Generator (TP-EGG). As shown in Figure~\ref{fig:model}, TP-EGG takes several seed predicates as input of the LM-based predicate generator to depict the domain of predicates and generate more in-domain predicates. With generated predicates, TP-EGG uses a novel transitivity-ensured edge selector by representing predicates as spheres in the vector space, to pick out the potential entailment relations among generated predicates. Then TP-EGG calculates the corresponding edge weights by the LM-based edge calculator. Our key insight is that by re-modeling the predicate extraction process as a generation process, we can leverage the underlying knowledge about natural language inference inside the LMs to avoid the data sparsity issues of extractive methods. By choosing appropriate seed predicates and setting the parameters of TP-EGG, one can generate EGs containing knowledge from a specific domain in arbitrary scales to fit the downstream requirement, without limitations from 
the uncontrollable distribution in domain-independent corpora. Since almost all the EG construction modules in TP-EGG is controlled by pre-trained LMs, the output EGs can be seen as explicit representations of the knowledge in LMs and used in downstream tasks, such as RTE in our experiments. 

In a word, our contributions can be summarized as follows: 
    (1) We propose a novel generative EG construction method to alleviate the data sparsity issues on generated EGs and avoid the reliance on corpora preparation in traditional EG methods;
    (2) We propose a new method to evaluate the quality of EGs in downstream tasks such as RTE;
    (3) Our TP-EGG outperforms strong baselines with significant improvement on benchmark datasets, and we show that generation-based EGs methods can alleviate the predicate sparsity by leveraging pre-trained LMs as predicate generators.

\section{Related Work}

\label{related-work}

Previous EG construction methods construct feature representations for typed predicates, weighted by counts or Pointwise Mutual Information~\citep{berant-etal-2015-efficient}, and compute the distribution similarity guided by DIH. For a predicate pair, different similarities are calculated, such as cosine similarity, Lin~\citep{lin-1998-automatic-retrieval}, Weed~\citep{weeds-weir-2003-general}, and Balanced Inclusion~\citep{szpektor-dagan-2008-learning}. Markov chain of predicate-argument transition~\citep{hosseini-etal-2019-duality} and temporal information from extracted corpora~\citep{guillou-etal-2020-incorporating} are also used in EGs construction. These methods independently calculate the entailment relations for each pair, called \textbf{local} methods. Besides, \textbf{global} constraints are used to detect new entailment relations beyond local relations. The transitivity in EGs, which means $a$ entails $b$ and $b$ entails $c$ indicate $a$ entails $c$ for three predicates $a,b$ and $c$, is the most widely used in previous works as hard constraints~\citep{berant-etal-2011-global,berant-etal-2015-efficient} or soft loss functions~\citep{hosseini-etal-2018-learning,chen-etal-2022-entailment}. The weight similarity constraints between different typed EGs and similar predicates are also taken into consideration~\citep{hosseini-etal-2018-learning}.

As one of the most important areas of NLP, text generation, or Natural Language Generation (NLG), has also been advanced by the surgent development of pre-trained LMs. BART~\citep{lewis-etal-2020-bart} uses encoder-decoder transformer architecture to re-correct the corrupted data in pre-training phase; GPT-3~\citep{NEURIPS2020_1457c0d6} uses transformer decoder to achieve in-context learning with massive multi-task unsupervised data. T5~\citep{10.5555/3455716.3455856} unifies different tasks into natural language prefixes and solves them by text generation.

Pre-trained LMs are also applied in recent EG methods. CNCE~\citep{hosseini-etal-2021-open-domain} initializes the contextualized embeddings of entity-relation triplets by BERT~\citep{devlin-etal-2019-bert} and uses random walk to get the entailment probability; EGT2~\citep{chen-etal-2022-entailment} fine-tunes a pattern-adapted LM on the predicate sentences and re-calculates high-quality edge weights for global constraints; \citet{McKenna2022SmoothingEG} applies RoBERTa~\citep{DBLP:journals/corr/abs-1907-11692} as predicate encoder and matches missing predicates in EGs with K-Nearest Neighbor algorithm to alleviate the predicate sparsity. As far as we are concerned, our method is the first attempt to use generative LM in EG construction and directly generate EGs without the distributional features from large corpora.

\section{Our Approach}

EGs store predicates as nodes and entailment relations between them as edges in graph structures. Following previous EG methods~\citep{hosseini-etal-2018-learning,hosseini-etal-2019-duality,chen-etal-2022-entailment}, we use the neo-Davisonian semantic form of binary relation \citep{parsons1990events} to indicate typed predicates, whose types are defined by the combination of argument types. Predicate $p$ connecting two arguments $a_1,a_2$ with types $t_1,t_2$ can be represented as $p=(w_1.i_1,w_2.i_2,t_1,t_2)$, where $w_j$ is the center relation tokens (and perhaps prepositions) about $a_j$, and $i_j$ is corresponding argument order of $a_j$ in $w_j$. For example, the event "\textit{The government is elected in 1910 and adored by natives}" contains two predicates $(elect.2,elect.in.2, government,time)$ and $(adore.1,adore.2, person,government)$. We denote $P$ as the collection of all typed predicates, $T$ as the collection of all argument types, and $\tau_1,\tau_2:P\to T$ as type indicator functions, where $\tau_1(p)=t_1$ and $\tau_2(p)=t_2$ for any predicate $p=(w_1.i_1,w_2.i_2,t_1,t_2)$.

We formally define that a typed entailment graph $G(t_1,t_2)$$=<$$P(t_1,t_2), E(t_1,t_2)$$>$ includes the collection of typed predicates $P(t_1,t_2)=\{p|(\tau_1(p),\tau_2(p))\in\{(t_1,t_2),(t_2,t_1)\}\}$, and the directional weighted edge set $E(t_1,t_2)$, which can be represented as an adjacent matrix $W(t_1,t_2)\in [0,1]^{|P(t_1,t_2)|\times |P(t_1,t_2)|}$. 
For those $G(t_1,t_2)$ whose $t_1\ne t_2$, the order of types $t1,t2$ is naturally determined. When $t_1=t_2=t$, argument types are ordered such that $G(t,t)$ can determine the order of types like "Thing A" and "Thing B" to distinguish predicates like "\emph{Thing A eat Thing B}" and "\emph{Thing B eat Thing A}". This order obviously affect the meaning of predicates, as “Thing A eats Thing B” entails “Thing B is eaten by Thing A”, but “Thing eats Thing” is doubtful to entail “Thing is eaten by Thing”.

\subsection{Predicate Generation}
In order to avoid the predicate sparsity issue in a given corpus, TP-EGG uses a predicate generator $\mathcal{G}$ to generate novel in-domain predicates. $\mathcal{G}$ takes a set of seed predicates $P_{seed}\subset P(t_1,t_2)$ as input and outputs a set of generated predicates $P_\mathcal{G}$, where $P_{seed}$ are expected to contain the domain knowledge of required EGs and $P_\mathcal{G}$ should be semantically related to $P_{seed}$ in varying degrees.

Our $\mathcal{G}$ is designed to be based on generative LMs, thus the input predicates $p\in P_{seed}$ should be converted into natural language forms to fit in the LMs. We use \citet{chen-etal-2022-entailment}'s sentence generator $S$ to convert predicate $p$ into its corresponding sentence $S(p)$. For example, $p=(elect.2,elect.in.2,government,time)$ will be converted into \emph{Government A is elected in Time B}. With converted sentences, generator $\mathcal{G}$ uses a generative LM, T5-large~\citep{10.5555/3455716.3455856} in our experiments, to generate new sentences and then re-converts them into generated predicates by a sentence-predicate mapping function $S^{-1}$~(details in Appendix \ref{sec:s-1}). Starting from the seed sentences $S_0=\{S(p)|p\in P_{seed}\}$, the generative LM outputs sentences $S_1$ for the next step, and $S_1$ is used to generate $S_2$ and so on, while $S^{-1}$ is used to re-convert $S_i$ to $P_i=S^{-1}(S_i)$ for every step. The generation process continues until the union of seed predicates and generated predicates $P'_i=P_{seed}\cup P_1...\cup P_i$ is equal to $P'_{i-1}$ or its size $|P'_i|$ exceeds a pre-defined scale parameter $K_{p}$.

To use T5-large as the generation component, we need to design an input template to generate new sentences. For sentence $s\in S_i$, the input template will be constructed like:
\centerline{$s$, which entails that $t_1$ A \emph{<extra\_id\_0>} $t_2$ B.}
\centerline{$s$, which entails that $t_2$ B \emph{<extra\_id\_0>} $t_1$ A.}
where \emph{<extra\_id\_0>} is the special token representing the generating location of the T5-large output. The max length of stripped output sequence $s'$ is limited to 5, and the new predicate $p'$ is produced by $S^{-1}($"$t_1$ A $s'$ $t_2$ B."$)$ or $S^{-1}($"$t_2$ B $s'$ $t_1$ A."$)$ correspondingly. For each $s$, T5-large uses beam-search algorithm with beam size $K_{beam}$ to find top-$K_{sent}$ output sequences $s'$ with highest probabilities. 

To ensure the quality of generated predicates and filter noisy ones, only those predicates which are generated by T5-large from at least two different predicates in $P'_{i-1}$ could be included in $P_i$. Algorithm~\ref{alg::pg} depicts how predicate generator $\mathcal{G}$ works (more details and examples in Appendix~\ref{sec:pgen_example}). 

\begin{algorithm}[t!]
% \begin{breakablealgorithm}
\small
\caption{The predicate generator $\mathcal{G}$.} 
\label{alg::pg} 
\begin{algorithmic}[1] 
\Require 
A set of seed predicates $P_{seed}$, sentence generator $S$, parameter $K_{beam},K_{sent},K_p$
\Ensure 
A set of generated predicates $P_{\mathcal{G}}$
\State{$P_{once}=\{\}$}
\State{$i=0,P_0=P'_0=P_{seed}$}
\While{$|P'_i|\le K_p$}
\State{$S_i=\{S(p)|p\in P_i\}$}
\State{$P_{i+1}=\{\}$}
\For{$s\in S_i$}
\State{$S^{g}=T5(s,K_{beam},K_{sent})$}
\State{$P^{g}=Set({S^{-1}(s^g)|s^g\in S^g})$}
\State{$P^{g}=P^{g}-P'_i$}
\State{$P_{i+1}$.update($P^{g}\cap P_{once}$)}
\State{$P_{once}=P_{once}$ XOR $P_g$}
\EndFor
\State{$P_{i+1}=P_{i+1}-P'_{i}$}
\State{$P'_{i+1}=P'_{i}\cup P_{i+1}$}
\If{$P'_{i+1}=P'_{i}$}
\Return{$P_{\mathcal{G}}=P'_i$}
\EndIf
\State{$i=i+1$}
\EndWhile
\Return{$P_{\mathcal{G}}=P'_i$}
\end{algorithmic} 
% \end{breakablealgorithm}
\end{algorithm}

\subsection{Edge Selection}

After generating new predicates $P(t_1,t_2)=P_{\mathcal{G}}$, TP-EGG constructs $G(t_1,t_2)$ by generating weighted edge set $E(t_1,t_2)$. As TP-EGG does not use large corpora to calculate distributional features regarding context coherence, we need to determine which predicate pairs could be potential entailment relations for later calculation. Regarding ALL pairs as candidates is a simple solution, but when $P(t_1,t_2)$ scales up, calculating all $|P|^2$ pairs will be unacceptably expensive as we intend to adopt an LM-based edge weight calculator, which only takes one pair as input at a time. Therefore, we require an effective edge selector $\mathcal{M}$ to select potential pairs $E'\subset P(t_1,t_2)\times P(t_1,t_2)$ with acceptable computational overhead, where $|E'|$ should be equal to a given parameter $K_{edge}$.

Calculating embeddings for each predicate and quickly getting similarities between all pairs in $P(t_1,t_2)$ perform worse than pair-wise LMs with cross attention in general, but are good enough as the edge selector to maintain high-quality pairs in high ranking. Inspired by \citet{10.1145/3106426.3106465}, we represent predicate $p$ as a sphere in the vector space. TP-EGG uses BERT-base~\citep{devlin-etal-2019-bert} to calculate embedding vector $v_p$ for every predicate $p$ based on $S(p)$, and represents $p$ as a sphere $\odot_p$ in a vector space with center $c_p$ and radius $r_p$:

\begin{equation}
\begin{aligned}
\label{sphere}
    v_p&=BERT(S(p))\in R^{d_v},\\
    c_p&=f_c(v_p)\in R^{d_c},\\
    r_p&=f^+(f_r(v_p))\in R_+.
\end{aligned}
\end{equation}
where $f_c,f_r$ are two-layer trainable neural networks, $d_v,d_r$ are corresponding vector dimensions, $f^+(x)\in\{\exp(x),x^2\}$ ensures the positive radius. By representing $p$ as a sphere, we expect that when $p$ entails $q$, $\odot_q$ should enclose $\odot_p$, as all points in $\odot_p$ are also included in $\odot_q$. Under such assumption, the transitivity referred in Section~\ref{related-work} is naturally satisfied as $\odot_a\subset\odot_b\subset\odot_c$. The overlapping ratio between spheres can be seen as the entailment probability $Pr(p\to q)$, and we simplify the calculation of sphere overlapping to diameter overlapping along the straight line between two centers:

\begin{equation}
\begin{aligned}
\label{p_overlap}
    d_{pq}&=||c_p-c_q||_2\\
    Pr(p\to q)&=\left\{\begin{array}{ll}
    0&,r_q\le d_{pq}-r_p\\
    1&,r_q\ge d_{pq}+r_p\\
    \frac{r_p+r_q-d_{pq}}{2r_p}&, otherwise\\
    \end{array}
    \right.
\end{aligned}
\end{equation}

\citet{chen-etal-2022-entailment} defines soft transitivity as $Pr(a\to b)Pr(b\to c)\le Pr(a\to c)$ for all predicate pairs above a threshold. Similar in spirit, our simplified sphere-based probability holds transitivity in part:

\newtheorem{Theorem}{Theorem}
\begin{Theorem} \label{theorem:1}
Given a threshold $\epsilon\in(0,1)$, $\forall a,b,c$ where $Pr(a\to b)>\epsilon$ and $Pr(b\to c)>\epsilon$, we have $Pr(a\to c)>\epsilon-(1-\epsilon)\frac{r_b}{r_a}$.
\end{Theorem}
We give its proof in Appendix~\ref{sec:appendix_proof}. Noted that while $\epsilon$ is close to $1$, the right part $\epsilon-(1-\epsilon)\frac{r_b}{r_a}$ will be nearly equal to $\epsilon$. As we use this probability in edge selection, higher $Pr(a\to b)$ and $Pr(b\to c)$ will naturally ensure the appearance of $(a,c)$ in final entailment relations, without the disturbance from low-confident edges. As $Pr(p\to q)$ is constant when $r_q\le d_{pq}-r_p$ or $r_q\ge d_{pq}+r_p$, its gradient becomes zero which makes it untrainable. Therefore, we smooth it with order-preserving Sigmoid function and interpolation, and finally get the selected edge set for $G(t_1,t_2)$:

\begin{equation}
\begin{aligned}
    \mathcal{M}(p,q)&=\sigma(\frac{2r_q-2d_{pq}}{r_p}),\\
    E(t_1,t_2)&=\{top K_{edge}(\mathcal{M}(p,q))|p,q\in V(t_1,t_2)\}
\end{aligned}
\end{equation}
where $\sigma$ is Sigmoid function $\sigma(x)=1/(1+e^x)$. A geometrical illustration presenting how the selector $\mathcal{M}$ works can be found in Appendix~\ref{sec:illu_m}.

\subsection{Edge Weight Calculation}

With the selected edge set $E(t_1,t_2)\subset P(t_1,t_2)\times P(t_1,t_2)$, TP-EGG calculates the edge weight $W_{p,q}$ for each predicate pairs $(p,q)$ individually in the adjacent matrix $W(t_1,t_2)$. Inspired by \citet{chen-etal-2022-entailment}, as the distributional features of generated predicates are unavailable for TP-EGG, we re-implement their local entailment calculator $\mathcal{W}$ to obtain the entailment edge weight $W_{p,q}$. $\mathcal{W}$ is based on DeBERTa~\citep{DBLP:journals/corr/abs-2006-03654,DBLP:journals/corr/abs-2111-09543} and fine-tuned to adapt to the sentence patterns generated by $S$. The entailment-oriented LM will produce three scores, corresponding to entailment~(E), neutral~(N) and contradiction~(C) respectively, for each sentence pair. The score of entailment class is used as the entailment edge weight in our EGs:

\begin{equation}
    W_{p,q}=\mathcal{W}(p,q)=\frac{exp(LM({\rm E}|p,q))}{\sum_{r\in\{{\rm E,N,C}\}}exp(LM(r|p,q))}
\end{equation}
where $LM(r|p,q)$ is the score of class $r$ After calculating all predicate pairs $(p,q)\in E(t_1,t_2)$ by the LM-based calculator $\mathcal{W}$, TP-EGG completes the adjacent matrix $W(t_1,t_2)$, and consequently constructs $G(t_1,t_2)$, as shown in Figure~\ref{fig:model}.

\section{Experimental Setup}

\paragraph{Datasets.}

Following previous works~\citep{hosseini-etal-2018-learning,hosseini-etal-2019-duality,hosseini-etal-2021-open-domain,chen-etal-2022-entailment}, we include Levy/Holt Dataset~\citep{levy-dagan-2016-annotating, holt2019probabilistic} and Berant Dataset~\citep{berant-etal-2011-global} into EG evaluation datasets. Besides, we reorganize the SherLIiC Dataset~\citep{schmitt-schutze-2019-sherliic}, a dataset for Lexical Inference in Context~(LIiC), into an EG benchmark. We further re-annotate conflicting pairs in Levy/Holt, referred as Levy/Holt-r Dataset. Dataset statistics are shown in Table~\ref{tab:data}. More details can be found in Appendix~\ref{sec:levyr}.

\begin{table}[]
    \centering
    \setlength{\tabcolsep}{3pt}
    \begin{tabular}{lccccc}
    \hline
    Name & Valid & Test & Total & \#Pos/\#Neg \\
    \hline
    Levy/Holt & 5,486 & 12,921 & 18,407 & 0.270 \\
    Levy/Holt-r & 5,450 & 12,817 & 18,267 & 0.261 \\
    Berant & - & 39,012 & 39,012 & 0.096 \\
    SherLIiC & 996 & 2,989 & 3,985 & 0.498 \\
    \hline
    \end{tabular}
    \caption{The dataset statistics.}
    \label{tab:data}
\end{table}

\paragraph{Metrics.}
Following previous works, we evaluate TP-EGG on the test datasets by calculating the area under the curves (AUC) of Precision-Recall Curve (PRC) for ROC curve.\footnote{We have found that the evaluation scripts written by \citet{hosseini-etal-2018-learning} do not connect the curve with (1,0) and (0,1) point correctly, which wrongly decreases the performance. We fix and use the scripts to generate results in this paper.} The evaluated EGs are used to match the predicate pairs in datasets and return the entailment scores. Noted that our generated predicates might be semantically same with required ones but have different forms, like \emph{(use.2,use.in.2,thing,event)} and \emph{(be.1,be.used.in.2,thing,event)} are both reasonable for \emph{"Thing A is used in Event B"} while our $S^{-1}$ generates the first one. Hence we relax the predicate matching standard in evaluation from exactly matching to sentence matching, i.e., $S(p)=S(p')$ rather than $p=p'$. This modification has nearly no effect on previous extraction-based EGs, but can better evaluate generative methods.

\begin{table*}[]
    \centering
\begin{tabular}{lcccccccc}
\hline
\multirow{2}{*}{\textbf{Methods}} & \multicolumn{2}{c}{L/H} & \multicolumn{2}{c}{L/H-r} & \multicolumn{2}{c}{Berant} & \multicolumn{2}{c}{SherLIiC}\\
% \hline
\cline{2-9}
 & PRC & ROC & PRC & ROC & PRC & ROC & PRC & ROC\\
\hline
BInc \citep{szpektor-dagan-2008-learning} & .262 & .632 & .254 & .632 & .242 & .676 & .170 & .605\\
\citet{hosseini-etal-2018-learning} & .271 & .638 & .254 & .637 & .268 & .682 & .184 & .611 \\
\citet{hosseini-etal-2019-duality} & .275 & .640 & .270 & .640 & .213 & .678 & .148 & .566\\
CNCE \citep{hosseini-etal-2021-open-domain} & .301 & .643 & .300 & .645 & .269 & .705 & .233 & .602\\
EGT2-Local \citep{chen-etal-2022-entailment} & .453 & .733 & .447 & .732 & .562 & .779 & .385 & .665 \\
 - w/ $L_3$ global & .477 & .755 & .478 & .756 & .583 & .780 & .391 & \bf .705\\
\hline
TP-EGG$_{L/H-r}$ & .543 & .755 & .527 & .748 & \cellcolor[gray]{0.9} .633 & \cellcolor[gray]{0.9} .780 & \cellcolor[gray]{0.9} .175 & \cellcolor[gray]{0.9} .606 \\
 - w/ EGT2-$L_1$ global & \bf .549 & \bf .778 & \bf .532 & \bf .773 & \bf \cellcolor[gray]{0.9} .637 & \cellcolor[gray]{0.9} \bf .822 & \cellcolor[gray]{0.9} .184 & \cellcolor[gray]{0.9} .615 \\
TP-EGG$_{SherLIiC}$ & \cellcolor[gray]{0.9} .263 & \cellcolor[gray]{0.9} .589 & \cellcolor[gray]{0.9} .261 & \cellcolor[gray]{0.9} .588 & \cellcolor[gray]{0.9} .171 & \cellcolor[gray]{0.9} .642 & \bf .394 & .669\\
 - w/ EGT2-$L_1$ global & \cellcolor[gray]{0.9} .264 & \cellcolor[gray]{0.9} .616 & \cellcolor[gray]{0.9} .261 & \cellcolor[gray]{0.9} .616 & \cellcolor[gray]{0.9} .173 & \cellcolor[gray]{0.9} .658 & \bf .394 & .680 \\
\hline
\end{tabular}
    \caption{The main results for TP-EGG$_{L/H-r}$, TP-EGG$_{SherLIiC}$ and baselines on EG benchmark datasets. The best performances of each metric are \textbf{boldfaced}, and the out-domain results are with gray ground color.}
    \label{tab:main_result}
\end{table*}

\paragraph{Implementation Details.}
In experiments, TP-EGG uses BERT-base in $\mathcal{M}$ and T5-large in $\mathcal{G}$ implemented by the Hugging Face transformer library~\citep{wolf-etal-2020-transformers}\footnote{https://github.com/huggingface/transformers}, and DeBERTa re-implementation from \citet{chen-etal-2022-entailment} to fine-tune on MNLI and adapt to sentence pattern in $\mathcal{W}$. Taking both EG performance and computational overhead into account, we set $K_p=5\times 10^3$, $K_{edge}=2\times 10^7$, $K_{beam}=50$, $K_{sent}=50$, $d_r=16$, $d_v=768$. Discussion about $K_p$ and $K_{edge}$ can be found in Appendix~\ref{sec:scale}.

\label{sec:implement}

For EG generation, TP-EGG uses the predicates in validation set of Levy/Holt-r and SherLIiC Dataset respectively as the seed predicate $P_{seed}$. With different $P_{seed}$, we also only use corresponding validation set as the training data for all later modules to keep the EGs \emph{in-domain}, called TP-EGG$_{L/H-r}$ and TP-EGG$_{SherLIiC}$ respectively. 

Only positive pairs are used to generate the training inputs and outputs to fine-tune T5-large in the predicate generator $\mathcal{G}$ with learning rate $\alpha_\mathcal{G}=10^{-3}$. We use $f^+(x)=\exp(x)$ for TP-EGG$_{L/H-r}$ and $f^+(x)=x^2$ for TP-EGG$_{SherLIiC}$. The edge selector $\mathcal{M}$ is also trained by the validation predicate pairs, but the positive examples are repeat 5 times (for Levy/Holt-r) or 2 times (for SherLIiC) to alleviate the label imbalance in training. BERT-base parameters are trained with learning rate $\alpha_{\mathcal{M},1}=10^{-5}$, while other parameters, including $f_c$ and $f_r$, are trained with learning rate $\alpha_{\mathcal{M},2}=5\times 10^{-4}$. The edge weight calculator $\mathcal{W}$ is trained by the same method in \citet{chen-etal-2022-entailment}. 

All modules are trained by AdamW optimizer~\citep{loshchilov2018decoupled} with cross entropy loss function, and controlled by early-stop mechanism, which stops the training when performances (loss for $\mathcal{G}$ and $F_1$ for others) on validation set do not reach the highest in the last 10 epoches. It takes about 5-6 hours to train all modules in TP-EGG, and about 2-3 hours to generate a typed EG on GeForce RTX 3090. The three modules, $\mathcal{G},\mathcal{M}$ and $\mathcal{W}$, contain 738M, 109M and 139M parameters respectively.

To be comparable with previous works~\citep{hosseini-etal-2018-learning}, we apply their lemma-based heuristic on all datasets except SherLIiC, and their average backup strategy on all datasets.

\paragraph{Compared Methods}
We compare TP-EGG with the best local distributional feature, Balanced Inclusion or called BInc \citep{szpektor-dagan-2008-learning}, and existing state-of-the-art local and global EG construction methods, including \citet{hosseini-etal-2018-learning,hosseini-etal-2019-duality}, CNCE~\citep{hosseini-etal-2021-open-domain} and EGT2~\citep{chen-etal-2022-entailment}.

\paragraph{Downstream Task.}
Despite of evaluating on EG construction benchmarks, we adapt an LM-based three-way RTE framework into the EG evaluation testbed. For premise $pm$ and hypothesis $h$, RTE models take their concatenation $[pm;h]$ as inputs, and return three probability scores of three classes. In order to incorporate the knowledge in EGs into RTE models, we design the following architecture available to any LM-based RTE model: given $pm$ and $h$, we extract binary predicates from them, and try to match the predicates in our EGs. Each matched predicates $a$ in premise $pm$ will be replaced by its $K_{nbr}$ neighbors $b$ with highest weight $W_{ab}$. For $h$, the neighbors $b$ are with highest weight $W_{ba}$. Replaced sentences $pm_1,...,pm_j$ and $h_1,...,h_k$ for $pm$ and $h$ will be concatenated to represent the information from EGs in calculation: 
\begin{equation}
\begin{aligned}
    (s_{\mathrm{E}1},s_{\mathrm{N}1},s_{\mathrm{C}1})=\mathrm{Softmax}(LM_1([pm;h])),\\
    (s_{\mathrm{E}2},s_{\mathrm{N}2},s_{\mathrm{C}2})=\mathrm{Softmax}(LM_2([pm;\\pm_1;...;pm_j;h;h_1;...;h_k])),\\
    s_i=(s_{i1}+s_{i2})/2,\quad i\in \{\mathrm{E,N,C}\}.
\end{aligned}
\end{equation}
where $LM_1$ and $LM_2$ represent two different LMs followed by a linear layer respectively. As the additional calculation unfairly requires more parameters, we also consider the models with equal parameters but do not use the EGs, referred as \emph{NO-EG} setting, by inputting $[pm;h]$ into $LM_2$ directly. We use SNLI~\citep{bowman-etal-2015-large} and SciTail~\citep{Khot2018SciTaiLAT} as our RTE benchmark datasets. We use BERT-base and DeBERTa-base as the backbone, learning rate $\alpha_{RTE}=10^{-5}$, $K_{nbr}=5$ for SNLI and $K_{nbr}=3$ for SciTail.

\section{Results and Analysis}

\subsection{Main Results}

\begin{figure}[t!]
    \centering
    \includegraphics[width=\linewidth]{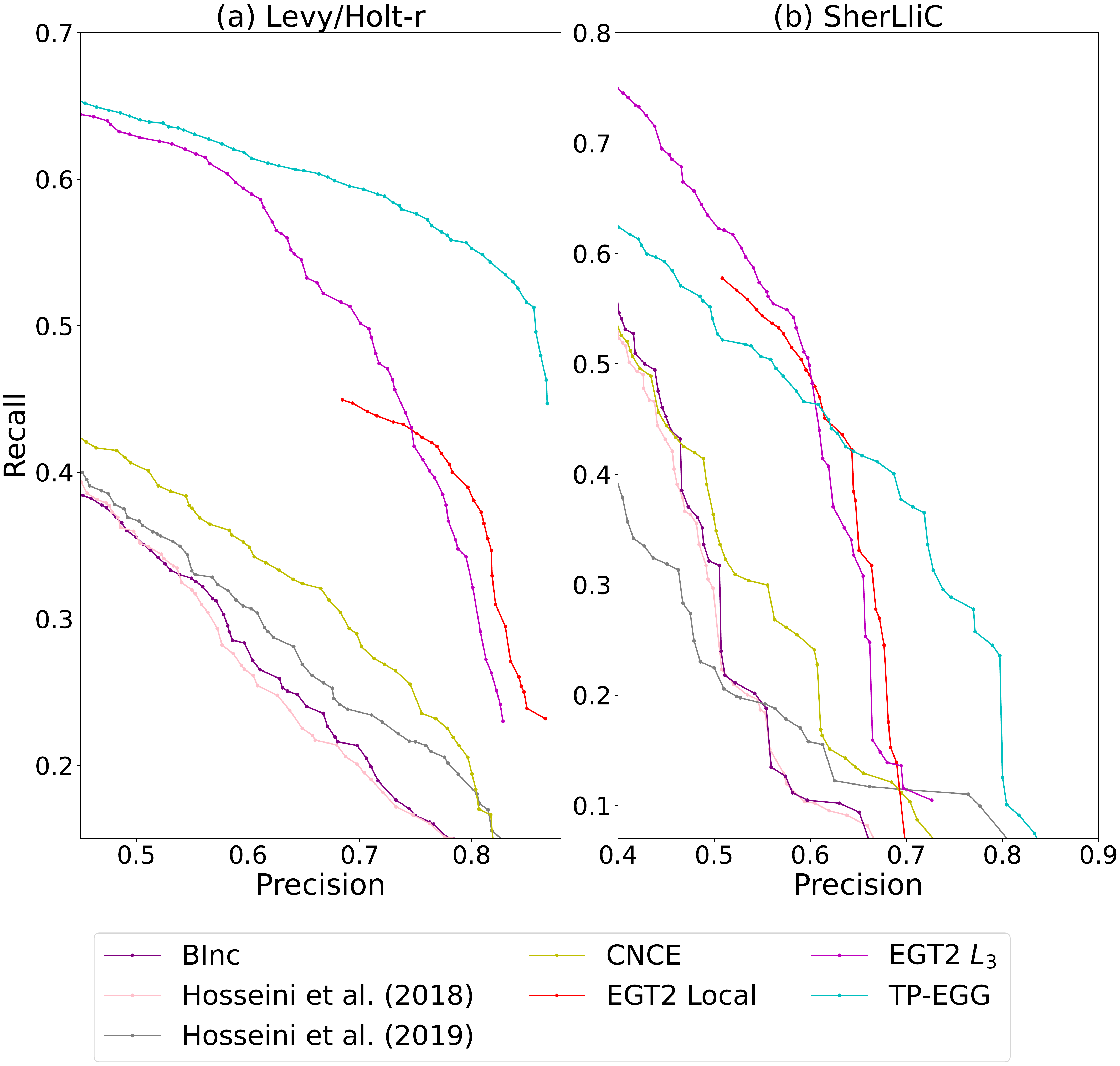}
    \caption{The Precision-Recall Curves of EGs on (a) Levy/Holt-r Dataset and (b) SherLIiC Dataset. For TP-EGG, the EGs are constructed with in-domain data.
    }
    \label{fig:curve}
\end{figure}

The performance of different EGs on benchmark datasets are shown in Table \ref{tab:main_result}, and the Precision-Recall Curves of EGs on Levy/Holt-r and Berant datasets are presented in Figure \ref{fig:curve}. Without using extracted features from large corpora, TP-EGG achieves significant improvement or at least reaches comparable performance with baselines for in-domain evaluations (L/H and L/H-r for TP-EGG$_{L/H-r}$ and SherLIiC for TP-EGG$_{SherLIiC}$). Interestingly, TP-EGG always performs better on the AUC of PRC, which indicates the strong ability of our generative methods to maintain impressive recall with high precision as shown in the curves. On Levy/Holt-r, TP-EGG$_{L/H-r}$ significantly outperforms all other extraction-based methods on precision>0.5, showing that with higher classification threshold, extraction-based methods fail to detect the entailment relations between rare predicates due to the sparsity issues, while generation-based TP-EGG successfully finds these relations by generating more predicates and correctly assigns high probabilities between them.

Noted that our TP-EGG is a local method, although certain global properties are ensured by our edge selector $\mathcal{M}$. We try to apply a state-of-the-art global method, EGT2-$L_1$~\citep{chen-etal-2022-entailment} on our local EGs\footnote{\citet{chen-etal-2022-entailment} reports that $L_3$ variant performs better on their local graphs, but we find $L_1$ is better on TP-EGG.}. As shown in the bottom of Table~\ref{tab:main_result}, the global method further improves the performance of TP-EGG, demonstrating the potential of our local EGs to continuously reducing the data sparsity with global EG learning methods. 

Although we have observed the significant improvement of evaluation metrics by TP-EGG, it is not clear enough to determine TP-EGG can alleviate the predicate sparsity to what extent. Therefore, we count the predicate pairs in Levy/Holt testset that exactly appeared as edges in EGs. We find that 6,873 pairs appear in TP-EGG$_{L/H-r}$, meanwhile 875 in EGT2-$L_3$. The far more appearance of in-domain predicates indicates the alleviation of predicate sparsity.

\begin{table}[]
    \centering
    \begin{tabular}{lcc}
    \hline
     \textbf{Methods} & PRC & ROC \\
    \hline
   BInc & .538 & .528\\
    \citet{hosseini-etal-2018-learning} & .535 & .529 \\
    \citet{hosseini-etal-2019-duality} & .554 & .556\\
    CNCE & .557 & .561\\
    EGT2-Local & .597 & .604\\
     - w/ EGT2-$L_3$ global & .626 & .644\\
    TP-EGG$_{L/H-r}$ & .609 & .596 \\
     - w/ EGT2-$L_1$ global & .636 & .633 \\
   \hline
    \end{tabular}
    \caption{Performance on the directional portion of Levy/Holt Dataset.}
    \label{tab:dire}
\end{table}

Previous works have claimed that LMs for entailments might be strong in undirectional paraphrasing, but weak in directional entailment recognizing~\citep{cabezudo2020natural,chen-etal-2022-entailment}. To check out the directional entailment ability of TP-EGG and other methods, we evaluate them on the directional portion\footnote{\url{https://github.com/mjhosseini/entgraph_eval/tree/master/LevyHoltDS}} of Levy/Holt Dataset as shown in Table~\ref{tab:dire}. The directional portion contains entailment pairs $(p,q)$ where $(p\to q)XOR(q\to p)$ is \emph{True}, and therefore symmetric models will have AUC<0.5. TP-EGG performs better than baselines on the directional portion, and the AUC far higher than 0.5 indicates its directional entailment ability. Global models perform better here, which is reasonable as global constraints are strongly related to the directional reasoning.

\subsection{Learning with Multiple Domains}

Although TP-EGG performs well on in-domain evaluation, the out-domain scenario is still hard, as the knowledge required for out-domain evaluation is inaccessible in all training and generation steps of TP-EGG. To check the impact of training data domains in different modules of TP-EGG, we use Levy/Holt-r and SherLIiC Dataset to produce seed predicates $P_{seed}$ and train different modules, including predicate generator $\mathcal{G}$, edge selector $\mathcal{M}$ and weight calculator $\mathcal{W}$, with different combinations of two datasets. As shown in Table \ref{tab:compo}, involving in-domain training data into more modules will lead to higher performance on corresponding dataset in general, which is in accordance with expectation.

\begin{table}[]
    \centering
    \begin{tabular}{clllcc}
    \hline
     & $P_{seed}$ & $\mathcal{G}$ & $\mathcal{M},\mathcal{W}$ & L/H-r & SLIC \\
    \hline
   \ding{172} & L/H-r & L/H-r & L/H-r & .527 & .175 \\ 
   \ding{173} & L/H-r & L/H-r & SLIC & .426 & .213 \\ 
   \ding{174} & L/H-r & SLIC & L/H-r & .411 & .323 \\ 
   \ding{175} & L/H-r & SLIC & SLIC & .312 & .384 \\ 
   \ding{176} & SLIC & L/H-r & L/H-r & .452 & .261 \\ 
   \ding{177} & SLIC & L/H-r & SLIC & .361 & .328 \\ 
   \ding{178} & SLIC & SLIC & L/H-r & .307 & .320 \\ 
   \ding{179} & SLIC & SLIC & SLIC & .261 & .392 \\ 
   \hline
    \end{tabular}
    \caption{Performance (AUC of PRC) on Levy/Holt-r and SherLIiC with different combinations of training data and modules. SLIC represents SherLIiC.}
    \label{tab:compo}
\end{table}

\begin{table}[]
    \centering
    \begin{tabular}{cllcc}
    \hline
    &$P_{seed}$ & $\mathcal{G},\mathcal{M},\mathcal{W}$ & L/H-r & SherLIiC \\
    \hline
   \ding{182}&L+S & L+S & .496 & .388 \\ 
   \ding{183}&L/H-r & L/H-r & .527 & .175 \\ 
   \ding{184}&L+S & L/H-r & .532 & .286 \\ 
   \ding{185}&L/H-r & L+S & .518 & .321 \\
   \ding{186}&SherLIiC & SherLIiC & .261 & .394 \\ 
   \ding{187}&L+S & SherLIiC & .322 & .416 \\ 
   \ding{188}&SherLIiC & L+S & .405 & .367 \\
   \hline
    \end{tabular}
    \caption{Performances (AUC of PRC) on Levy/Holt-r Dataset and SherLIiC Dataset of TP-EGG trained with merged multi-domain data.}
    \label{tab:rsjoin}
\end{table}

Interestingly, by comparing different combinations, we find that fine-tuning $\mathcal{G}$ with data from domains different with $P_{seed}$ will lead to better overall performance on two datasets. For example, row \ding{174} attains improvement about 0.15 on SherLIiC with dropping about 0.11 on Levy/Holt-r by changing training data of $\mathcal{G}$ from Levy/Holt-r (\ding{172}) to SherLIiC, and when $P_{seed}$ also changes to SherLIiC (\ding{178}), the performance on Levy/Holt-r is severely damaged without benefit to SherLIiC. Similar situation is also observed in row \ding{173},\ding{175} and \ding{179}. We assume that involving knowledge from different domains in predicate generation, i.e. $P_{seed}$ and $\mathcal{G}$, could alleviate the over-fitting by mixing two predicate domains and encouraging $\mathcal{G}$ to find more novel predicates to cover the gap between training and testing. Empirically, involving different data in $\mathcal{G}$ leads to the best performance among the modules.

Next, we study the effect of using merged validation sets of Levy/Holt-r and SherLIiC Dataset at different modules. The performance of TP-EGG trained with the merged data, referred as L+S, are shown in Table~\ref{tab:rsjoin}. While using merged data as $P_{seed}$ and also as training data for other modules (\ding{182}), TP-EGG reaches impressive performances on both datasets, which is not surprising, as both datasets are in-domain in this situation. 

Using merged dataset to train $\mathcal{G}, \mathcal{M}$ and $\mathcal{W}$ boosts out-domain performance with in-domain performance loss (comparing \ding{183} and \ding{185}, \ding{186} and \ding{188}). However, adding some out-domain predicates into $P_{seed}$ is surprisingly beneficial to the in-domain evaluation while improving out-domain generalization (comparing \ding{183} and \ding{184}, \ding{186} and \ding{187}). We attribute it to the diversity of generated predicates led by the newly incorporated seed predicates, which might not be generated with the in-domain seed predicates. The out-domain predicates help TP-EGG to find new predicates related to in-domain predicates as Algorithm~\ref{alg::pg} might tend to generate predicates from at least two predicates across two domains. Therefore, the predicate coverage over evaluation datasets can be increased.

\subsection{Results on RTE}

In downstream task evaluation, we use EGs generated by different methods to enhance LM-based RTE models, and report the results in Table~\ref{tab:nliresult}. Compared with CNCE and EGT2, our TP-EGG achieves better performance on two RTE datasets with both BERT$_{base}$ and DeBERTa$_{base}$ backbones. The performances of TP-EGG on DeBERTa$_{base}$ are significantly better than NO-EG (p<0.05). Noted that TP-EGG offers $pm_j,h_k$ for 4,600 sentences in SNLI testset, which is 5,596 for EGT2-$L_3$. Even with lower coverage over predicates in the dataset, TP-EGG supports RTE models with more high-quality entailment relations to generate $pm_j,h_k$ and improve the performance. On the other hand, the noisy entailment relations in CNCE and EGT2 perhaps misguide RTE models, thus lead to even worse results than \emph{NO-EG} in some cases.

\begin{table}[]
    \centering
    \setlength{\tabcolsep}{3pt}
    \begin{tabular}{llcc}
    \hline
         Model & EG & SNLI & SciTail \\
         \hline
         \multirow{5}{*}{BERT} & Original & 90.03$\pm$0.04 & 91.42$\pm$0.21 \\
          & NO-EG & 90.17$\pm$0.19 & 92.64$\pm$0.07 \\
          & CNCE & 90.10$\pm$0.19 & 92.15$\pm$0.98 \\
          & EGT2-$L_3$ & 90.08$\pm$0.05 & 92.35$\pm$0.05 \\
          & TP-EGG & \bf 90.28$\pm$0.22 & \bf 92.94$\pm$0.92 \\
         \hline
         \multirow{5}{*}{DeBERTa} & Original & 91.59$\pm$0.26 & 94.20$\pm$0.55\\
          & NO-EG & 91.69$\pm$0.03 & 94.62$\pm$0.23 \\
          & CNCE & 91.57$\pm$0.19 & 95.06$\pm$0.33 \\
          & EGT2-$L_3$ & 91.35$\pm$0.24 & 94.57$\pm$0.46 \\
          & TP-EGG & \bf 91.90$\pm$0.11 & \bf 95.19$\pm$0.20 \\
      \hline
    % \hline
    \end{tabular}
    \caption{Performances of RTE models supported with different EGs on RTE datasets (average over 3 runs). The best performances are \textbf{boldfaced}.}
    \label{tab:nliresult}
\end{table}

\subsection{Ablation Study}

We run the ablation experiments which directly use the original version of LMs in $\mathcal{G},\mathcal{M}$ and $\mathcal{W}$ without fine-tuning on EG benchmark datasets. For $\mathcal{M}$, as non-LM parameters are involved, we replace it with \emph{randomly} selecting $K_{edge}$ edges. As shown in Table~\ref{tab:ablation}, without fine-tuning $\mathcal{G}$ or $\mathcal{W}$, the performance on Levy/Holt-r suffers a significant drop (about 0.1), indicating the importance of fine-tuned modules for EG generation. The performance on SherLIiC also decreases severely without fine-tuning $\mathcal{G}$, as the fine-tuning step can improve the quality of generated predicates and cover more out-domain predicates. Fine-tuning $\mathcal{W}$ critically affects the result on Berant Dataset, which is compatible with the results in \citet{chen-etal-2022-entailment}, showing the importance of fine-tuning and pattern adaptation in weight calculation on this dataset. Fine-tuning $\mathcal{M}$ is mainly beneficial to SherLIiC by comparison. From the results, we can see that high quality predicate pair construction from $\mathcal{G}$ and $\mathcal{M}$ is more beneficial to out-domain evaluation, while the weight calculation from $\mathcal{W}$ plays a more important role for in-domain cases.

\begin{table}[]
    \centering
    \setlength{\tabcolsep}{4pt}
    \begin{tabular}{lccc}
    \hline
         Method & L/H-r & Berant & SherLIiC \\
         \hline
         TP-EGG$_{L/H-r}$ & .527 & .633 & .175 \\
          - w/o fine-tuning $\mathcal{G}$ & .422 & .508 & .132 \\
          - w/o training $\mathcal{M}$ & .518 & .615 & .152 \\
          - w/o fine-tuning $\mathcal{W}$ & .429 & .305 & .166 \\
      \hline
    \end{tabular}
    \caption{Experiment results of ablation study with different modules in TP-EGG.}
    \label{tab:ablation}
\end{table}

\section{Conclusions}

In this work, we propose a novel generative typed entailment graph construction method, called TP-EGG, with predicate generation, edge selection and calculation modules. TP-EGG takes several seed predicates as input to the predicate generator to find novel predicates, selects potential entailment predicate pairs as edges, and calculates the edge weights without distributional features. TP-EGG can construct high-quality EGs with flexible scales and avoid the data sparsity issues to some extent Experiments on EG benchmacks and RTE task show the significant improvement of TP-EGG over the state-of-the-art EG learning methods. We find that mixing data from different domains in different ways can improve the generalization of TP-EGG in varying degrees, and using out-domain data in predicate generation modules brings the most significant improvement.

\section*{Limitations}

First, as we do not rely on specific corpora and avoid the shortcomings of extractive methods, we also lose their advantages. The typed EGs generated by our TP-EGG is strongly related to the seed predicates and training data of generation modules, while extractive EGs can generate domain-independent EGs from large corpora and do not require supervised training data to a considerable degree. Second, the edge calculator $\mathcal{W}$ is time-consuming even we can control the scales of output EGs, as the edge num $|E(t_1,t_2)|$ will be relatively large for TP-EGG to generate powerful EGs. Furthermore, how to effectively select seed predicates still remains a difficult problem which has not been discussed thoroughly in this work by using the validation datasets. We assume that this problem could be solved by carefully confirming how the seed predicates represent corresponding domain knowledge and we leave it to future work.

\section*{Ethics Statement}

We re-annotate the Levy/Holt Dataset which is a publicly available dataset for entailment graph evaluation. Annotators receive a competitive pay of about 100 yuan per hour under the agreement of the institute, which is more than 4 times the local minimum wage. The annotation complies with the ACL Code of Ethics. The sentences used in annotation are generated from the original dataset and we do not incorporate external content into the sentences. However, there may still be sentences containing potentially improper content, which do not reflect the views or stances of the authors. The re-annotation results are confirmed by the majority voting of annotators, and may still contain natural errors. Further usage of the re-annotated dataset should be aware of the limitation and the authors are not responsible for any issues in further usage of this dataset.

\section*{Acknowledgements}

This work is supported in part by NSFC (62161160339). We would like to thank the anonymous reviewers for their helpful comments and suggestions.

\bibliography{anthology,custom}
\bibliographystyle{acl_natbib}

\appendix

\section{The Proof of Theorem 1}
\label{sec:appendix_proof}

\textbf{Theorem 1} \emph{Given a threshold $\epsilon\in(0,1)$, $\forall a,b,c$ where $Pr(a\to b)>\epsilon$ and $Pr(b\to c)>\epsilon$, we have $Pr(a\to c)>\epsilon-(1-\epsilon)\frac{r_b}{r_a}$.}

As $Pr(p\to q)=\frac{r_p+r_q-d_{pq}}{2r_p}$ holds when $d_{pq}-r_p<r_q<d_{pq}+r_p$, and $Pr(p\to q)=1\le \frac{r_p+r_q-d_{pq}}{2r_p}$ holds when $r_q\ge d_{pq}+r_p$, we have:
\begin{equation}
\begin{aligned}
    \frac{r_a+r_b-d_{ab}}{2r_a}\ge Pr(a\to b)>\epsilon\\
    \to d_{ab}<r_b+(1-2\epsilon) r_a.
\end{aligned}
\end{equation}
Similarly, for $b,c$:
\begin{equation}
    d_{bc}<r_c+(1-2\epsilon) r_b.
\end{equation}

For the case $Pr(a\to c)=1$, obviously the theorem holds for $\epsilon\in(0,1)$;

For the case $Pr(a\to c)=0$ or $Pr(a\to c)=\frac{r_p+r_q-d_{pq}}{2r_p}$, we have $Pr(a\to c)\ge \frac{r_p+r_q-d_{pq}}{2r_p}$ as $r_p+r_q-d_{pq}<0$ under $Pr(a\to c)=0$, and therefore:
\begin{equation}
\begin{aligned}
    & Pr(a\to b)\\
    \ge & \frac{r_a+r_c-d_{ac}}{2r_a}\\
    \ge & \frac{r_a+r_c-(d_{ab}+d_{bc})}{2r_a}\quad(d_{ac}\le d_{ab}+d_{bc} )\\
    > & \frac{r_a+r_c-(r_b+(1-2\epsilon)r_a+r_c+(1-2\epsilon)r_b)}{2r_a}\\
    = & \frac{\epsilon r_a+(\epsilon-1)r_b}{r_a}\\
    = & \epsilon+(\epsilon-1)\frac{r_b}{r_a}.
\end{aligned}
\end{equation}
Q.E.D.

\section{Geometrical Illustration of Edge Selector $\mathcal{M}$}
\label{sec:illu_m}

\begin{figure}[t!]
    \centering
    \includegraphics[width=0.95\linewidth]{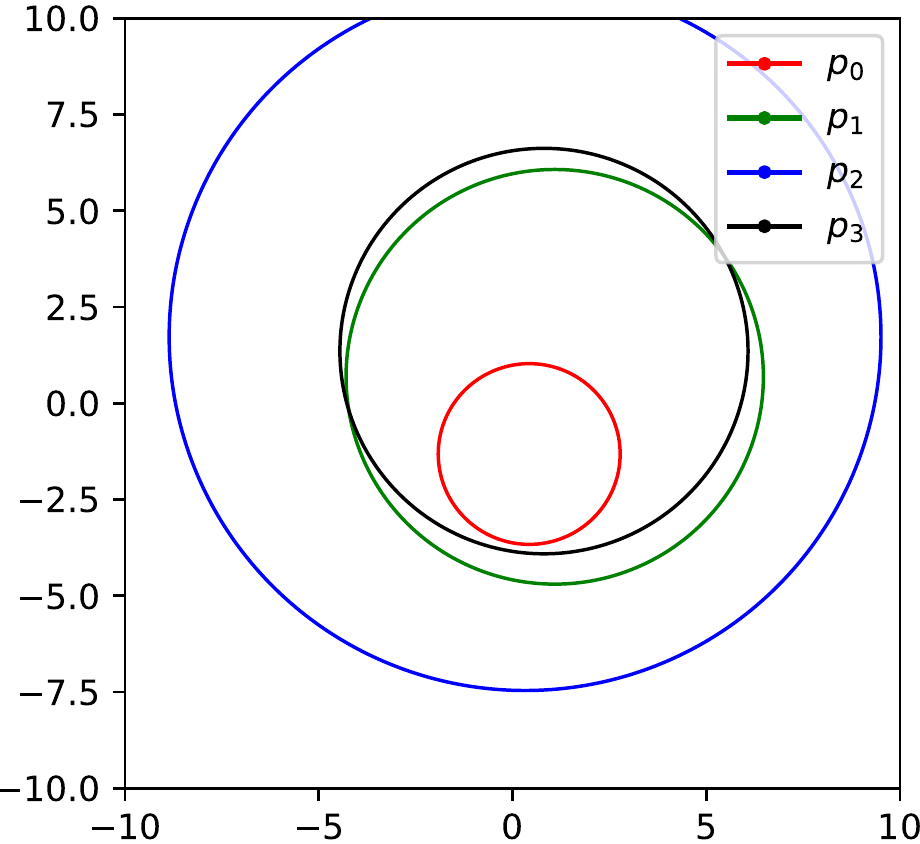}
    \caption{The visualized predicate spheres of four predicates.}
    \label{fig:illu_m}
\end{figure}

To understand how the edge selector $\mathcal{M}$ works more intuitively, we pick four predicate sentences from Levy/Holt Dataset and visualize their corresponding spheres $\odot_p$ in Figure~\ref{fig:illu_m}:

\centerline{$p_0$: Living Thing A is imported from Location B.}
\centerline{$p_1$: Living Thing A is native to Location B.}
\centerline{$p_2$: Living Thing A is found in Location B.}
\centerline{$p_3$: Living Thing A is concentrated in Location B.}

The centers $c_p$ and radius $r_p$ are generated by $\mathcal{M}$ from our final TP-EGG model, while the dimension of $c_p$ are reduced to maintain the distances between them. Three entailment relations, $p_0\to p_1$, $p_1\to p_2$ and $p_3\to p_2$, are annotated in the dataset, and $p_0\to p_3$ is also plausible. In Figure~\ref{fig:illu_m}, the hypothesis spheres obviously enclose premise spheres, and the more generic a predicate is, the bigger its sphere becomes, which is consistent with our expectation about $\mathcal{M}$. With high directional overlapping, all of the four entailment relations will correctly appear in later weight calculation while low-confident inverse edges will be filtered out.

\section{The Sentence-Predicate Mapping Function $S^{-1}$}
\label{sec:s-1}

The sentence-predicate mapping function $S^{-1}$ used in predicate generation is described in Algorithm~\ref{alg::s-1}. Noted that $S^{-1}$ is a simplified approximation of the reverse function of sentence generator $S$ while different predicates might generate the same sentence by $S$. Therefore, $S^{-1}$ does not cover all possible predicates and sentences.

\begin{breakablealgorithm}
\small
\caption{The mapping function $S^{-1}$.} 
\label{alg::s-1} 
\begin{algorithmic}[1] 
\Require 
A generated sentence $s$.
\Ensure 
A predicate $p$, or $NULL$ indicating that $s$ is not a valid predicate sentence.
\State{Split the sentences into tokens $l$ and strip $t_1$ A, $t_2$ B}
\State{prefix=""}
\If{$|l|=0$}
\Return{NULL}
\EndIf
\If{\emph{not} or \emph{n't} in l[0]}
\State{prefix=$NEG$\quad // representing the negation}
\EndIf
\State{Remove the modal verbs in $l$}
\If{$l$ begins with \emph{have been} or \emph{has been}}
\State{l=l[1:]}
\EndIf
\If{$|l|>1$ and  $l[:2]$ is \emph{have+P.P.}}
\State{l=l[1:]}
\EndIf
\If{$|l|>2$ and the present tense of $l[:2]$ is \emph{have to}}
\State{l=l[2:]}
\EndIf
\If{$|l|=0$}
\Return{NULL}
\EndIf
\State{$i_{head}=0,i_{tail}=|l|-1$}
\While{$i_{head}\le i_{tail}$ and $l[i_{head}]$ is not a verb}
\State{$i_{head}=i_{head}+1$}
\EndWhile
\While{$i_{head}\le i_{tail}$ and $l[i_{tail}]$ is not a verb or a preposition}
\State{$i_{tail}=i_{tail}-1$}
\EndWhile
\If{$i_{head}>i_{tail}$}
\Return{NULL}
\EndIf
\State{$l'=l[i_{head}:i_{tail}+1]$\quad // cut the token between $i_{head}$ and $i_{tail}$}
\If{$l'[0:2]$ is a verb like \emph{be doing}}
\State{$l'=l'[1:]$}
\EndIf
\State{$t=lemmatize(l'[0])$}
\If{$t$ is \emph{be}}
\If{$|l'|=1$}
\Return{prefix+$(be.1,be.2,t_1,t_2)$}
\EndIf
\If{$l'[1]$ is not a preposition}
\If{$l'[1]$ is an adverb}
\State{$l'=l'[0:1]+l'[2:]$}
\EndIf
\If{$l'[1]$ is an adjective or a noun, and $l'[-1]$ is a preposition}
\State{$l'[1]=lemmatize(l'[1])$}
\Return{prefix+$(l'[1].1,l'[1:].2,t_1,t_2)$}
\EndIf
\If{$l'[1]$ is P.P. verb}
\State{$l'[1]=lemmatize(l'[1])$}
\If{$l'[-1]$ is a preposition}
\Return{prefix+$(l'[1].2,l'[1:].2,t_1,t_2)$}
\Else
\Return{prefix+$(l'[1].2,l'[1:].3,t_1,t_2)$}
\EndIf
\EndIf
\EndIf
\Return{NULL}
\EndIf
\State{$l'[0]=lemmatize(l'[0])$}
\If{$|l'|=1$}
\Return{prefix+$(l'[0].1,l'[0].2,t_1,t_2)$}
\EndIf
\If{$l'[-1]$ is a preposition}
\Return{prefix+$(l'[0].1,l'.2,t_1,t_2)$}
\EndIf
\Return{NULL}
\end{algorithmic} 
\end{breakablealgorithm}

\section{An Example of Generating Predicates from Seed Predicates}
\label{sec:pgen_example}

We show an example process of generating new predicates by the generator $\mathcal{G}$ of TP-EGG in Table~\ref{tab:pgen_example}. We set $P_{seed}=\{p_1,p_2,p_3\}$, $K_{beam}=K_{sent}=8$, $K_p=15$. The predicates repeating in current generation or appearing in previous stages, and sentences that cannot be resolved by $S^{-1}$ are omitted. Predicates generated from at least two different $s$ are in \realred{red}, and predicates appeared in generation of previous steps are in \realblue{blue}. According to Algorithm \ref{alg::pg}, only seed predicates and colored predicates will appear in final predicate set $P_2'$.

\section{Discussion about Graph Scales}
\label{sec:scale}

\begin{figure}[t!]
    \centering
    \includegraphics[width=0.95\linewidth]{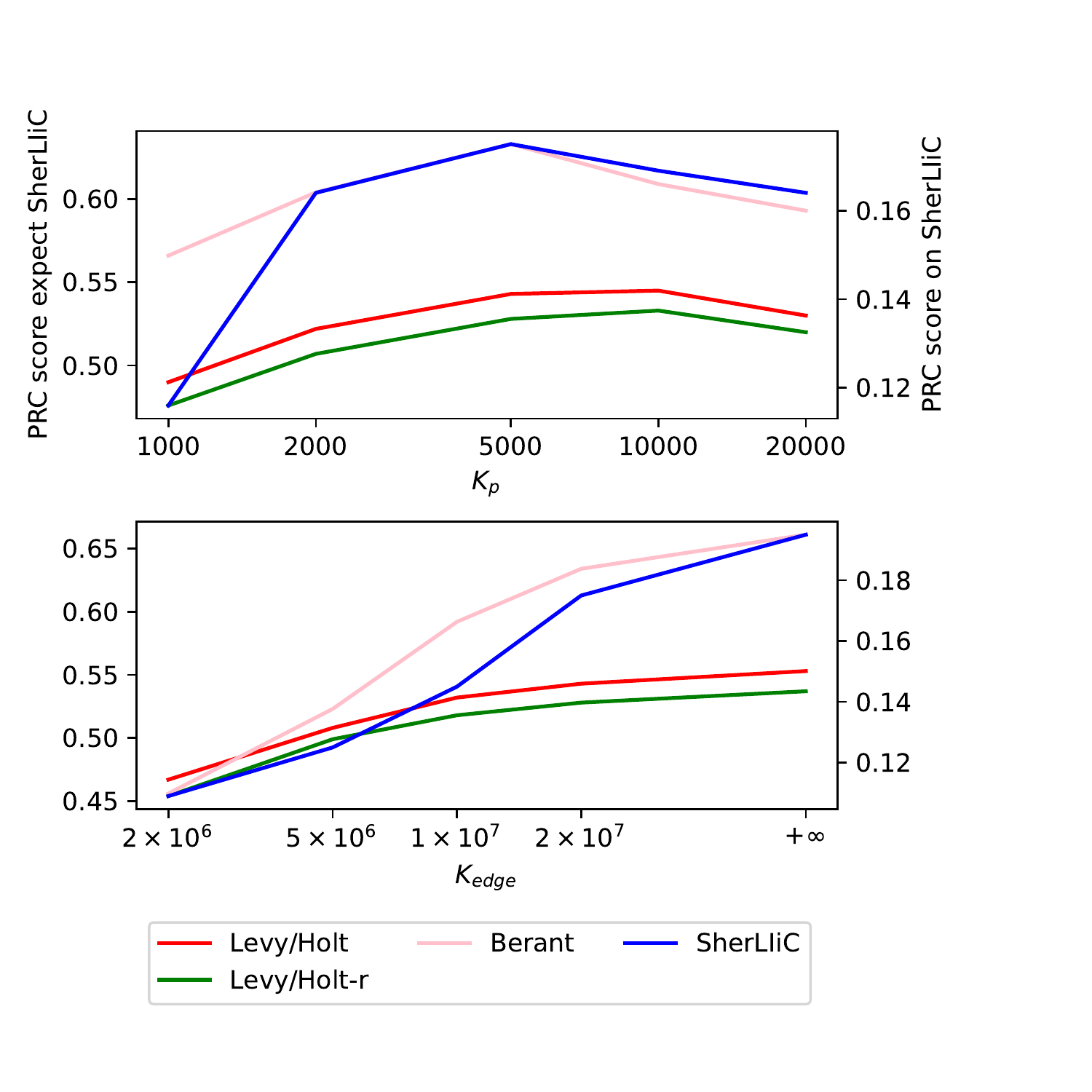}
    \caption{The performances on the evaluation datasets with different $K_p$ and $K_{edge}$ of TP-EGG$_{L/H-r}$. The y-axis of the curves on SherLIiC dataset is on the right side.
    }
    \label{fig:scale_curve}
\end{figure}

As referred in Section~\ref{sec:implement}, we set the number of predicates $K_p=5\times 10^3$ and edges $K_{edge}=2\times 10^7$, which determine the final scale of generated EGs. We report the performance of TP-EGG$_{L/H-r}$ on the evaluation datasets with different $K_p$ and $K_{edge}$ in Figure~\ref{fig:scale_curve}. Changing $K_p$ from $1\times 10^3$ to $2\times 10^4$, the overall performance is the best while $K_p=5\times 10^3$. We assume that lower $K_p$ might limit the coverage of predicate set, while higher $K_p$ makes the EGs more sparse and miss potential entailment relations. Noted that the computational overhead and space occupation is almost proportional to $K_{edge}$, setting $K_{edge}=+\infty$ to regard ALL pairs as candidates is impractical (the largest EG in TP-EGG$_{L/H-r}$ will contain $7\times 10^7$ edges). We find that $K_{edge}=2\times 10^7$ is able to reach the overall performances comparable with $K_{edge}=+\infty$ under our settings, while further decreasing $K_{edge}$ will significantly cut down the performances. To balance between the overall performance and computational overhead, we finally set $K_p=5\times 10^3$ and $K_{edge}=2\times 10^7$.

\section{Details about Datasets}
\label{sec:levyr}

\citet{levy-dagan-2016-annotating} uses questions and candidate answers with textual predicates to collect the entailment relations, and proposes a widely used EG evaluation dataset which is later re-annotated by \citet{holt2019probabilistic}, called Levy/Holt Dataset. For example, if the annotator figures out that \emph{"The government is adored by natives"} can be used to answer \emph{"Who recognize the government?"}, the dataset will indicate that \emph{"adore"} entails \emph{"recognize"} between type \emph{person} and \emph{government}. 
Levy/Holt Dataset contains 18,407 predicate pairs (14,491 negative and 3,916 positive). We use the 30\%/70\% splitting of validation/test set as \citet{hosseini-etal-2018-learning} in our experiments. 

However, because the QA annotation form incorporates additional information about entities related to the predicates, some consistent predicates pairs are annotated with different labels, and the transitivity is disobeyed between some predicate pairs. The inconsistent pairs are those $(a,b)$ which $(a,b,True)$ and $(a,b,False)$ both appear in the dataset. The transitivity-disobeying pairs are those $(a,b)$, $(b,c)$ and $(a,c)$ which $(a,b,True)$, $(b,c,True)$ and $(a,c,False)$ all appear. We find that there are 89 inconsistent pairs and 159 transitivity-disobeying pairs in Levy/Holt Dataset, and re-annotate these 248 pairs by five annotators with Fleiss' $\kappa=0.43$. After re-annotating, we get the final Levy/Holt-r Dataset with 14,490 negative and 3,777 positive pairs. 

\citet{berant-etal-2011-global} proposes an annotated entailment relation dataset, containing 3,427 positive and 35,585 negative examples, called Berant Dataset.

\citet{schmitt-schutze-2019-sherliic} extracts verbal relations from ClueWeb09~\citep{clueweb09} based on Freebase~\citep{10.1145/1376616.1376746} entities, and splits the extracted relations into typed one based on their most frequent Freebase types, which is naturally compatible to typed EG settings. We use their manually-labeled 1,325 positive and 2,660 negative examples in our EG benchmark, called SherLIiC Dataset. The dataset is split into 25\%(validation) and 75\%(test) in our experiments.

\onecolumn
\begin{longtable}{lp{13cm}}
    \hline
         Stage & Predicates and Sentences \\
         \hline
         $P_{seed}$($S_0$) & $p_1:$(adore.1, adore.2, person,government) (Person A adores Government B),\\
         &$p_2:$(recognize.1, recognize.2, person,government) (Person A recognizes Government B),\\ 
         &$p_3:$(know.1, know.2, person,government) (Person A knows Government B) \\
         \hline
         $S^g$($P^g$) & $p_1\to $\{Person A is identified with Government B. \realblue{(identify.2, identify.with.2, person,government)},\\
         &Person A is Government B. (be.1, be.2, person,government),\\
         &Government B is magnet for Person A. \realblue{(magnet.1, magnet.for.2, government,person)},\\
         &Government B is worshipped in Person A. (worship.2, worship.in.2, government,person),\\
         &Government B is drawn to Person A. \realred{(draw.2, draw.to.2, government,person)},\\
         &Person A is devoted to Government B. (devote.2, devote.to.2, person,government),\\
         &Person A is associated with Government B. \realred{(associate.2, associate.with.2, person,government)},\\
         &Government B is magnet of Person A. \realblue{(magnet.1, magnet.of.2, government,person)}\} \\
         &$p_2\to $\{Government B is family of Person A. (family.1, family.of.2, government,person), \\
         &Government B is associated with Person A. \realred{(associate.2, associate.with.2, government,person)},\\
         &Person A identifies with Government B. \realred{(identify.1, identify.with.2, person,government)},\\
         &Government B is drawn to Person A. \realred{(draw.2, draw.to.2, government,person)},\\
         &Person A is associated with Government B. \realred{(associate.2, associate.with.2, person,government)}, \\
         &Person A identifies with Government B. \realred{(identify.1, identify.with.2, person,government)},\\ 
         &Person A is connected with Government B. \realred{(connect.2, connect.with.2, person,government)},\\
         &Government B wants Person A. \realblue{(want.1, want.2, government,person)}\}\\
         &$p_3\to $\{Government B is associated with Person A. \realred{(associate.2, associate.with.2, government,person)}, \\
         &Person A identifies with Government B. \realred{(identify.1, identify.with.2, person,government)},\\ 
         &Government B awards Person A. \realblue{(award.1, award.2, government,person)},\\
         &Government B is drawn to Person A. \realred{(draw.2, draw.to.2, government,person)},\\
         &Person A embodies Government B. \realblue{(embody.1, embody.2, person,government)},\\ 
         &Person A is associated with Government B. \realred{(associate.2, associate.with.2, person,government)},\\
         &Person A is connected with Government B. \realred{(connect.2, connect.with.2, person,government)},\\
         & Government B is enemy of Person B. (enemy.1, enemy.of.2, government,person)\}\\
        \hline
         $P_1$ & $p_4:$(associate.2, associate.with.2, person,government)\\
         & $p_5:$(identify.1, identify.with.2, person,government)\\
         & $p_6:$(connect.2, connect.with.2, person,government)\\
         & $p_7:$(draw.2, draw.to.2, government,person)\\
         & $p_8:$(associate.2, associate.with.2, government,person)\\
         \hline
         $S^g$($P^g$) & $p_4\to $\{Person A is identified with Government B. \realblue{(identify.2, identify.with.2, person,government)},\\
         &Government B awards Person A. \realblue{(award.1, award.2, government,person)},\\
         &Person A practices Government B. \realred{(practice.1, practice.2, person,government)},\\
         &Government B is gravitate towards Person B. \realred{(be.1, be.gravitate.towards.2, government,person)},\\
         &Government B is sought after by Person A. (seek.2, seek.after.by.2, government,person),\}\\
         & $p_5\to $\{Government B issues call for Person A. \realred{(issue.1, issue.call.for.2, government,person)},\\
         &Person A declares Government B. (declare.1, declare.2, person,government), \\
         &Person A embodies Government B. \realblue{(embody.1, embody.2, person,government)},\\ 
         &Person A declares war on Government B. (declare.1, declare.war.on.2, person,government)\}\\
         & $p_6\to $ \{Person A is identified with Government B. \realblue{(identify.2, identify.with.2, person,government)},\\
         &Government B is after Person A. (be.1, be.after.2, government,person),\\
         &Government B issues call for Person A. \realred{(issue.1, issue.call.for.2, government,person)},\\
         &Government B is identified with Person A.  (identify.2, identify.with.2, government,person),\\
         &Person A practices Government B. \realred{(practice.1, practice.2, person,government)},\\
         &Person A embodies Government B. \realblue{(embody.1, embody.2, person,government)}\}\\
         & $p_7\to $\{Person A submits Government B. (submit.1, submit.2, person,government),\\
         &Government B is attracted to Person A. (attract.2, attract.to.2, government,person),\\
         &Government B is magnet for Person A. \realblue{(magnet.1, magnet.for.2, government,person)},\\
         &Person A believes in Government B> (believe.1, believe.in.2, person,government),\\
         &Government B is magnet of Person A. \realblue{(magnet.1, magnet.of.2, government,person)}\}\\
         & $p_8\to $\{Person A is identified with Government B. \realblue{(identify.2, identify.with.2, person,government)},\\
         &Person A preaches Government B. (preach.1, preach.2, person,government),\\
         &Government B issues call for Person A. \realred{(issue.1, issue.call.for.2, government,person)},\\
         &Person A practices Government B. \realred{(practice.1, practice.2, person,government)},\\
         &Person A demands Government B. (demand.1, demand.2, person,government),\\
         &Government B is gravitate towards Person B. \realred{(be.1, be.gravitate.towards.2, government,person)},\\
         &Government B wants Person A. \realblue{(want.1, want.2, government,person)}\} \\
         \hline
         $P_2$ & $p_9:$(identify.2, identify.with.2, person,government)\\
         & $p_{10}:$(magnet.1, magnet.for.2, government,person)\\
         & $p_{11}:$(issue.1, issue.call.for.2, government,person)\\
         & $p_{12}:$(award.1, award.2, government,person)\\
         & $p_{13}:$(practice.1, practice.2, person,government)\\
         & $p_{14}:$(embody.1, embody.2, person,government)\\
         & $p_{15}:$(be.1, be.gravitate.towards.2, government,person)\\
         & $p_{16}:$(want.1, want.2, government,person)\\
         & $p_{17}:$(magnet.1, magnet.of.2, government,person)\\
         \hline
         $P'_2$ & Return $p_1,...,p_{17}$ \\
      \hline
    % \end{tabular}
    \caption{An example of generating predicates $P_i'$ from $P_{seed}$.}\\
    \label{tab:pgen_example}\\
\end{longtable}

\end{document}